\documentclass[letterpaper]{article} 
\usepackage{aaai2027}  

\usepackage[hyphens]{url}  
\usepackage{graphicx} 
\usepackage{natbib}  
\usepackage{caption} 
\usepackage{algorithm}
\usepackage{algorithmic}
\usepackage{amsmath,amssymb,amsfonts}

\usepackage{newfloat}
\usepackage{listings}
\DeclareCaptionStyle{ruled}{labelfont=normalfont,labelsep=colon,strut=off} 
\floatstyle{ruled}
\newfloat{listing}{tb}{lst}{}
\floatname{listing}{Listing}

\usepackage{booktabs}
\usepackage{multirow}

\title{GeoGAT: Bidirectional Temporal Sampling Meets Hierarchical Graph Attention for Global Video Geo-localization}

\author{
	Junchao Cui\textsuperscript{\rm 1},
	Xuanzi Ma\textsuperscript{\rm 1},
	Wenqi Shi\textsuperscript{\rm 1},
	Hangyu Li\textsuperscript{\rm 1},
	Biru Zhu\textsuperscript{\rm 1},
	Chong Fu\textsuperscript{\rm 2},
	Xiangyang Luo\textsuperscript{\rm 1} \footnote{Corresponding author : {luoxy\_ieu@sina.com}}
}
\affiliations{
	\textsuperscript{\rm 1}Information Engineering University, Zhengzhou, China\\
	\textsuperscript{\rm 2}Zhengzhou University, Zhengzhou, China\\
}

\begin{document}
	\nocopyright
	\maketitle

\begin{abstract}
	Global video geo-localization aims to infer the geographic location of a video worldwide, evaluating performance across four geographic hierarchies: city, state/province, country, and continent. Existing methods typically employ one-way uniform sampling to process video frames and train independent classifiers for each hierarchy, which leads to the loss of key geographic cues and prediction conflicts between hierarchies, especially for complex multi-shot edited videos. To address these limitations, we propose GeoGAT, which integrates bidirectional temporal sampling with graph attention networks (GATs). Specifically, GeoGAT extracts forward and offset-reversed frame sequences to construct complementary spatiotemporal features. These fused features are then fed into a predefined geographical hierarchy graph, where GATs perform structure-aware message passing, while a dual-constraint mechanism prunes predictions to eliminate cross-hierarchy conflicts. We construct GeoGAT10k, comprising 9,720 multi-shot edited videos from 166 cities worldwide, specifically to benchmark generalization ability on complex video structures. Experimental results on CityGuessr68k and GeoGAT10k demonstrate that GeoGAT eliminates hierarchical conflicts entirely and achieves state-of-the-art performance across all four geographic hierarchies. On CityGuessr68k, GeoGAT outperforms the strongest baseline, evaluated under both classification and retrieval protocols, by 2.6 percentage points at the city level. On the more challenging GeoGAT10k with multi-shot edited videos, the accuracy improvement exceeds 24 percentage points, validating strong generalization to complex real-world scenarios.
\end{abstract}

\section{Introduction}
\label{Introduction} 
Global video geo-localization aims to infer the city-level geographic location of a video captured anywhere on Earth~\cite{1}. Unlike tasks that estimate precise coordinates from individual static images~\cite{2,3,4,5,6}, continuous camera movement renders the representation of an entire video by a single coordinate inappropriate, owing to the inherent spatial-temporal dynamics. With the explosive growth of video-sharing platforms such as TikTok and YouTube~\cite{7,8}, video volume has increased dramatically. This surge makes accurate global localization of video content increasingly critical for applications such as content verification, misinformation detection, and location-aware recommendation.

Current research on video geo-localization remains in its preliminary stages. Existing methods focus on specific regions~\cite{10,11,12}, such as individual cities~\cite{CITY,CITY2} or tourist attractions~\cite{CITY3}, using retrieval-based strategies. They first construct a reference database of geotagged images (typically ground-view and corresponding satellite views), then match video keyframes against database images. Early studies employed CNN~\cite{CNN} for cross-view matching~\cite{13,Cross-1,Cross-2}, while later work used VGG~\cite{VGG} and ViT~\cite{14,34} to enhance accuracy. However, these methods fundamentally belong to fine-grained localization paradigms, applicable only to limited predefined locations, and cannot scale to global coverage.

\begin{figure}[t]
	\centering
	{\includegraphics[width=1\linewidth,trim=0bp 0bp 0bp 0bp,clip=true]{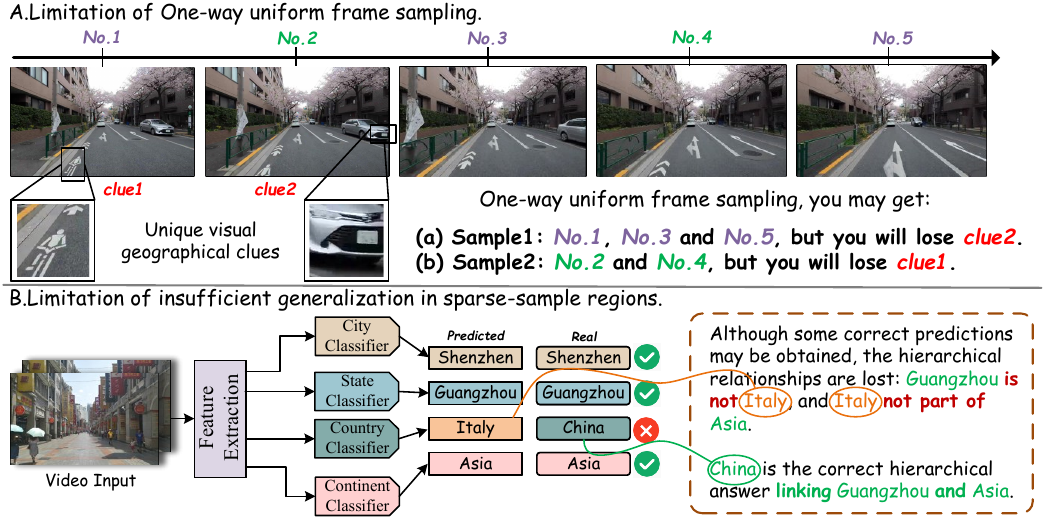}} 
	\caption{Research motivation.}
	\label{fig-1} 
\end{figure}

Recently, Kulkarni et al.~\cite{1} proposed CityGuessr, which formalizes global video geo-localization as a hierarchical classification problem. Their approach adopts VideoMAE~\cite{29} as the video encoder and employs four independent linear heads corresponding to the four geographic hierarchies. Auxiliary mechanisms, including soft-scene labels and text label alignment, further enhance feature representation, and the method achieves initial success on the accompanying CityGuessr68k benchmark~\cite{1}. Nevertheless, the core architecture still relies on uniform frame sampling and independent classifiers, which are unable to fully exploit sparse yet critical geographic cues. Moreover, hierarchical consistency constraints are applied only at inference time rather than during training, leaving hierarchical conflicts unresolved in the learning process. Fig.~\ref{fig-1} summarizes the limitations of existing methods.

Geographic cues in videos are sparse and unevenly distributed: landmarks, road signs, or textual identifiers often appear in only a few frames, and fixed-interval one-way sampling easily misses them amid redundant content. Moreover, geographic labels obey strict inclusion constraints (city $\in$ state $\in$ country $\in$ continent), yet existing methods train independent classifiers per hierarchy without structural constraints during training; soft remedies such as text label alignment cannot fundamentally eliminate the resulting cross-hierarchy conflicts.

To address these issues, we propose GeoGAT, which predicts labels at all four hierarchies for each video through three synergistic components. First, bidirectional temporal sampling extracts forward and offset-reversed frame sequences and fuses them at the feature level, recovering geographic cues missed by one-way uniform sampling. Second, a GAT-based hierarchy classifier models the predefined geographic hierarchy as a directed graph and performs structure-aware message passing in place of independent linear classifiers. Third, a dual-constraint mechanism couples soft parent-child correlation during training with hard geographic validity enforcement during inference, eliminating cross-granularity conflicts that message passing alone cannot resolve.

GeoGAT is trained to maximize the geographic information retained in the video representation, instantiated through two complementary variational bounds on the mutual information (Sec.~\ref{sec:loss}). To the best of our knowledge, this is the first work to introduce mutual information maximization and GATs into global video geo-localization. The existing benchmark CityGuessr68k~\cite{1} contains only single-shot continuous videos, inadequately evaluating generalization on complex video structures. We therefore construct and release GeoGAT10k, comprising 9,720 multi-shot edited videos from 166 cities worldwide, aiming to provide richer data support for future research. Our contributions can be summarized as follows:

\begin{itemize}
	\item We propose GeoGAT for global video geo-localization with two key components:  (1) bidirectional temporal sampling for complementary spatiotemporal features, and (2) a GAT-based hierarchical classifier with dual constraints to resolve cross-granularity prediction conflicts.
	\item We introduce GeoGAT10k, a dataset of 9,720 multi-shot edited videos from 166 cities worldwide, evaluating model generalization on complex video structures.
	\item We conduct extensive experiments on CityGuessr68k and GeoGAT10k, where GeoGAT achieves state-of-the-art performance across all four geographic hierarchies.
\end{itemize}

\section{Related Work}
\label{Related Work}
\subsection{Paradigms of Video Geo-localization}
\label{21}

Video geo-localization can be divided into two mainstream paradigms: retrieval-based~\cite{Cross-3,Cross-4} and classification-based~\cite{1}. The retrieval-based paradigm relies on large-scale reference databases, matching video frames with geotagged images for fine-grained localization in local regions~\cite{GeoCLIP}. VidTAG~\cite{kulkarni2026vidtag} formulates video geo-localization as cross-modal frame-to-GPS retrieval via contrastive learning. While achieving high accuracy, retrieval-based methods are limited by database construction costs and coverage constraints, making global scaling impractical. The classification-based paradigm formulates geo-localization as multi-classification, either through grid-based partitioning or administrative divisions, to address global-scale challenges. This paradigm first achieves success in image-based geo-localization~\cite{DualGeo} and is subsequently extended to video~\cite{1}, discretizing geographic space into manageable categories.

\subsection{Retrieval-based Region-level Paradigm}
\label{22}
Existing video geo-localization works primarily focus on fine-grained localization in local regions via cross-view retrieval. This paradigm treats video frames as queries and builds reference databases from satellite imagery for cross-view matching. Representative works include GTFL~\cite{12}, GAMa~\cite{11} with contrastive learning and 3D/2D CNN~\cite{CNN} encoders, and SeqGeo~\cite{10} for limited field of view. The effectiveness of such methods in local regions stems from two factors: limited scope enables high accuracy with small database overhead~\cite{16,12,18}, and the ready availability of satellite imagery supports database construction~\cite{19,20}. However, they inherently face scalability issues: database costs increase dramatically with region expansion, and they are only applicable to predefined local areas.

\subsection{Classification-based Global Paradigm}
\label{23}
Extending geo-localization to the global scale presents challenges of broad coverage, numerous categories, and high computational overhead. In the image domain, classification has become the dominant approach, with substantial research~\cite{21,22,23,24} providing methodological foundations for video tasks. Weyand et al.~\cite{3} first formulate global image geo-localization as classification with PlaNet, partitioning the Earth's surface into fixed grids. Subsequent works include CPlaNet~\cite{4} for fine-grained outputs, multi-level classification with kernel density estimation~\cite{25}, ISNs for scene-specific networks~\cite{5}, HRNet-based semantic segmentation~\cite{26} by Pramanick et al.~\cite{6}, and GeoDecoder with cross-attention~\cite{27}. Collectively, these works establish the effectiveness of classification for global-scale localization. Inspired by these, Kulkarni et al.~\cite{1} first formulate global video geo-localization as classification with CityGuessr, employing four independent linear heads with auxiliary mechanisms such as soft-scene labels and text label alignment. They also open-source CityGuessr68k, the first large-scale global video geo-localization benchmark. We construct GeoGAT10k to evaluate generalization under complex video structures, offering richer data support.

\section{Proposed Method}
\label{sec:method}

\begin{figure*}[t]
	\centering
	{\includegraphics[width=1\linewidth,trim=0bp 0bp 0bp 0bp,clip=true]{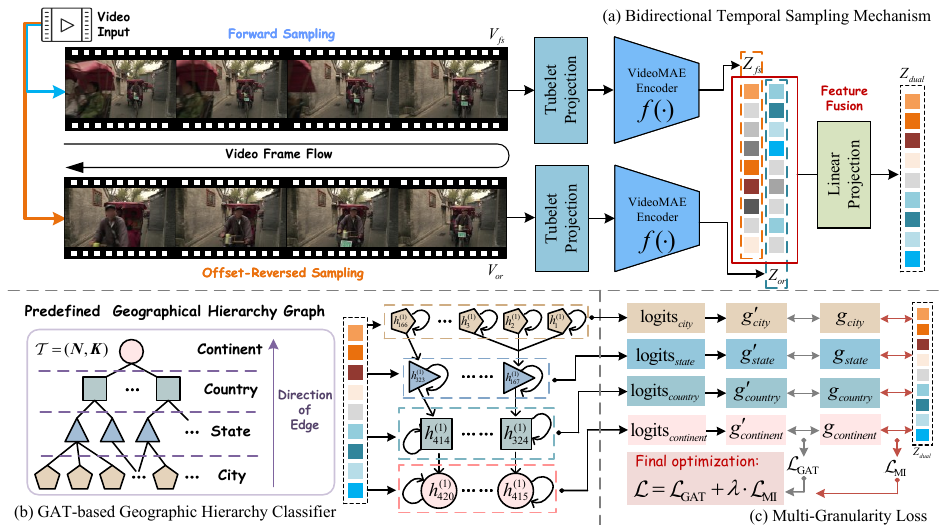}} 
	\caption{Overview of GeoGAT. \textbf{(a)}~Bidirectional Temporal Sampling extracts complementary features from forward and offset-reversed frame sequences. \textbf{(b)}~GAT-based Hierarchy Classifier propagates information over a directed geographic graph to model cross-granularity dependencies. \textbf{(c)}~Multi-granularity Loss enforces geographic consistency through mutual information and hierarchical constraints.}
	\label{fig:2} 
\end{figure*}

Existing video geo-localization methods suffer from two key limitations: \textit{(i)}~unidirectional uniform sampling misses sparse yet critical geographic cues buried in long videos, and \textit{(ii)}~independent classifiers ignore the inherent geographic hierarchy and cause conflicts across granularities. To address these issues, we propose GeoGAT, which couples bidirectional temporal sampling with GATs. As shown in Fig.~\ref{fig:2}, GeoGAT extracts complementary spatiotemporal features through dual sampling streams and then performs structure-aware message passing over a geographic hierarchy graph to resolve hierarchical conflicts at the architectural level.

\subsection{Problem Formalization}
\label{sec:problem}

Global video geo-localization aims to predict the capture location of a video $V = \{v_1, v_2, \dots, v_T\}$ containing $T$ frames. The target is a hierarchical geographic label $G = (g_{\text{city}}, g_{\text{state}}, g_{\text{country}}, g_{\text{continent}})$. Only a subset of frames contain discriminative geographic cues (\emph{e.g.}, landmarks, vegetation, and signage); the rest carry location-irrelevant dynamics (\emph{e.g.}, pedestrian motion and illumination variation). The core challenge lies in extracting maximal geographic information from limited and noisy observations.

We formulate this task as maximizing the mutual information $I(Z; G)$ between a $d$-dimensional video representation $Z = f(V)$ and the geographic label $G$. Drawing upon the information bottleneck principle~\cite{28}, the objective seeks an optimal compressed representation that preserves discriminative geographic signals while suppressing irrelevant content. To approach this information-theoretic optimum, we propose GeoGAT with two core steps:

\textbf{Step 1:} Bidirectional temporal sampling produces two complementary frame subsets $V_{\text{fs}}$ and $V_{\text{or}}$. A VideoMAE encoder~\cite{29} extracts features $Z_{\text{fs}}$ and $Z_{\text{or}}$, which are fused into a joint representation $Z_{\text{dual}}$ (Sec.~\ref{sec:sampling}).

\textbf{Step 2:} $Z_{\text{dual}}$ is projected into a predefined geographic hierarchy graph $\mathcal{T}$ to initialize node features. GAT layers then perform structure-aware message passing to align video embeddings with geographic node embeddings, yielding predictions for all four hierarchies (Sec.~\ref{sec:classifier}).

A multi-granularity loss (Sec.~\ref{sec:loss}) jointly supervises the optimization. Concretely, we optimize $I(Z; G)$ through two complementary variational bounds: the geo-localization loss (Eq.~\ref{eq:lgat}) maximizes the Barber--Agakov lower bound~\cite{barber2003algorithm}, $I(Z;G) \geq H(G) + \mathbb{E}[\log q_\theta(G \mid Z)]$, where $q_\theta$ is the predictive distribution of the hierarchy classifier and $H(G)$ is constant; the mutual information loss (Eq.~\ref{eq:infonce}) maximizes a contrastive lower bound on the same quantity~\cite{Oord2018Representation}. Both terms thus optimize $I(Z_{\text{dual}}; G)$ from complementary directions, making the information-theoretic objective and the practical training loss tightly coupled by construction.

\subsection{Bidirectional Temporal Sampling}
\label{sec:sampling}

Unidirectional uniform sampling with fixed intervals tends to miss sparse geographic cues while accumulating location-irrelevant dynamic noise, which limits the geographic information extractable from the video. To enrich the extracted signal, we propose a bidirectional sampling mechanism that constructs two complementary views: a forward sequence $V_{\text{fs}}$ and an offset-reversed sequence $V_{\text{or}}$.

Given a video with $T$ frames and a target sample size $m$:

\vspace{2pt}
\noindent\textbf{Forward Sampling.} Starting from the first frame, uniformly sample $m$ frames:
\begin{equation}
	U_{\text{fs}} = \left\{ \left\lfloor \tfrac{i \cdot T}{m} \right\rfloor \right\}_{i=0}^{m-1}, \quad V_{\text{fs}} = \{ v_i \mid i \in U_{\text{fs}} \}.
\end{equation}

\noindent\textbf{Offset-Reversed Sampling.} Introduce an offset $p$ to avoid overlap with the forward set, then reverse the order to break temporal directionality:
\begin{equation}
	U_{\text{or}}^{\text{raw}} = \left\{ \left\lfloor p + \tfrac{i \cdot T}{m} \right\rfloor \right\}_{i=0}^{m-1}, \quad U_{\text{or}} = \text{Reverse}(U_{\text{or}}^{\text{raw}}),
\end{equation}
with corresponding frames $V_{\text{or}} = \{ v_i \mid i \in U_{\text{or}} \}$.

Both sequences are encoded by a VideoMAE model~\cite{29} pre-trained on Kinetics-400~\cite{30}: $Z_{\text{fs}} = f(V_{\text{fs}})$ and $Z_{\text{or}} = f(V_{\text{or}})$. Concatenating and projecting these features yields $Z_{\text{dual}}$. Although both sequences are drawn from the same video, their disjoint temporal support and reversed ordering expose complementary geographic evidence. By the chain rule of mutual information,
\begin{equation}
	I([Z_{\text{fs}}, Z_{\text{or}}]; G) = I(Z_{\text{fs}}; G) + I(Z_{\text{or}}; G \mid Z_{\text{fs}}),
\end{equation}
where the second term quantifies the incremental geographic information contributed by the offset-reversed branch. Whenever the offset-reversed view captures discriminative cues missed by the forward view, which complementary temporal coverage promotes, this term is positive, yielding a representation richer in geographic signal than that of either branch alone.

\subsection{GAT-based Geographic Hierarchy Classifier}
\label{sec:classifier}

Linear classifiers~\cite{1,3,4,5,6} predict each geographic hierarchy independently, overlooking the structured constraint that city $\in$ state $\in$ country $\in$ continent. This causes \textit{(i)}~cross-granularity prediction conflicts and \textit{(ii)}~poor generalization in regions with sparse training data. We instead model the geographic space as a directed hierarchy graph $\mathcal{T} = (\mathbf{N}, \mathbf{K})$ and perform structure-aware message passing through GATv2~\cite{32}.

The node set $\mathbf{N}$ contains all geographic labels across four hierarchies, with nodes at each level distinguished by fixed offsets. The edge set $\mathbf{K}$ encodes parent-child ``belongs-to'' relationships, forming a forest of hierarchical trees that represents global geographic topology.

For each video, $Z_{\text{dual}}$ is projected into the graph embedding space to initialize node features. Each node $n \in \mathbf{N}$ also receives a learnable embedding $h_n$, initialized with decreasing variance from city to continent levels. This stabilizes higher-level embeddings against cascading errors while preserving flexibility for lower-level nodes.

Information propagates through $L$ GATv2 layers, each of which employs multi-head self-attention over neighboring nodes together with residual connections and layer normalization. The dynamic attention mechanism of GATv2~\cite{32} captures complex inter-node dependencies more flexibly than the static attention of GAT~\cite{31}. After $L$ layers:
\begin{equation}
	H^{(L)} = \text{GATv2}^L(\dots \text{GATv2}^2(\text{GATv2}^1(H^{(0)}, \mathbf{K})) \dots),
\end{equation}
where $H^{(0)}$ denotes the initial node embedding matrix.

For prediction, we compute the cosine similarity between $\ell_2$-normalized $Z_{\text{dual}}$ and the refined node embeddings $H^{(L)}$, scaled by a learnable temperature $\tau$ for each hierarchy $k \in \{\text{city}, \text{state}, \text{country}, \text{continent}\}$:
\begin{equation}
	\text{logits}_k = \exp(\tau) \cdot \text{sim}\!\left(\ell_2(Z_{\text{dual}}),\, H^{(L)}_k\right).
	\label{eq:logits}
\end{equation}

To enforce hierarchical consistency, we apply two constraints on the logits:

\noindent\textbf{Soft constraint.} Child node logits are augmented with weighted parent logits, so high child confidence drives up the parent score. Node embeddings are initialized with decreasing variance from city to continent (0.01, 0.009, 0.008, 0.005), which stabilizes upper-level predictions against cascading errors.

\noindent\textbf{Hard constraint.} At inference, a parent-child lookup table restricts the valid prediction space per granularity, enforcing the geographic rule city $\in$ state $\in$ country $\in$ continent. Logits are then sliced by node offsets to produce the four hierarchy-specific distributions.

\subsection{Multi-Granularity Loss}
\label{sec:loss}

The training objective combines two complementary losses: $\mathcal{L}_{\text{MI}}$ maximizes mutual information between the video representation and geographic labels, and $\mathcal{L}_{\text{GAT}}$ directly supervises hierarchical predictions.

\noindent\textbf{Mutual Information Loss.} We employ InfoNCE~\cite{Oord2018Representation} as a lower bound on $I(Z_{\text{dual}}; G)$. For hierarchy $k$, the positive pair $(Z_{\text{dual}}, g_k)$ is contrasted against the node embeddings of other labels within the same batch $\mathcal{B}$, where $g_k$ denotes the refined embedding of the ground-truth node at hierarchy $k$, taken from $H^{(L)}$ (Eq.~\ref{eq:logits}):
\begin{equation}
	\mathcal{L}_{\text{InfoNCE}}^{(k)} = -\log \frac{\exp\!\big(\text{sim}(Z_{\text{dual}}, g_k) / \tau\big)}{\sum_{j \in \mathcal{B}} \exp\!\big(\text{sim}(Z_{\text{dual}}, g_{k,j}) / \tau\big)}.
	\label{eq:infonce}
\end{equation}
The total mutual information loss is averaged over all four hierarchies:
\begin{equation}
	\mathcal{L}_{\text{MI}} = \frac{1}{4} \sum_{k} \mathcal{L}_{\text{InfoNCE}}^{(k)}.
\end{equation}

\noindent\textbf{Geo-localization Loss.} Cross-entropy loss directly supervises the prediction at each hierarchy:
\begin{equation}
	\mathcal{L}_{\text{GAT}} = \sum_{k} \text{CE}\!\left(\sigma(\text{logits}_k),\, g_k\right),
	\label{eq:lgat}
\end{equation}
where $\sigma(\cdot)$ denotes the softmax function.

\vspace{2pt}
\noindent\textbf{Final Loss.} The combined objective is:
\begin{equation}
	\mathcal{L} = \mathcal{L}_{\text{GAT}} + \lambda \cdot \mathcal{L}_{\text{MI}},
\end{equation}
where $\lambda$ balances the two terms. This joint objective drives the model to learn discriminative geographic representations while the hierarchy graph propagates structural knowledge across granularities, which improves robustness in sparse-sample regions and eliminates cross-granularity conflicts.

\section{Evaluation}
\label{Evaluation} 

\subsection{Dataset and Metrics}
\label{5.1}

\begin{figure}[t]
	\centering
	{\includegraphics[width=1\linewidth,trim=0bp 0bp 0bp 0bp,clip=true]{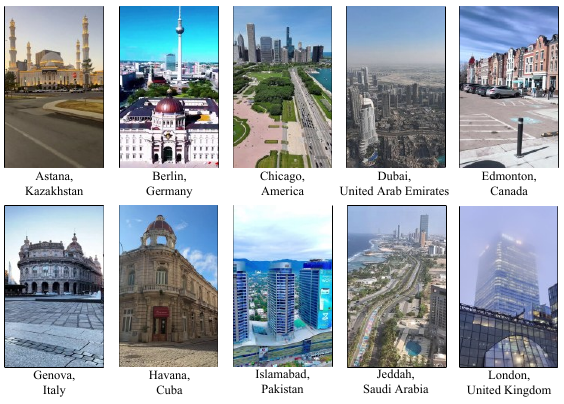}} 
	\caption{Video frame samples from 10 different countries in the GeoGAT10k dataset.}
	\label{fig-3} 
	
\end{figure}

\textbf{{GeoGAT10k dataset.}} To address the predominance of single-shot videos in existing datasets, we construct and release GeoGAT10k. All videos are collected and used with explicit permission from the original authors. Fig.~\ref{fig-3} shows example frames from various geographic locations. Each video is manually annotated with four geographic hierarchies to ensure label reliability. More details about GeoGAT10k is shown in the supplementary material.

\begin{figure}[t]
	\centering
	{\includegraphics[width=1\linewidth,trim=0bp 0bp 0bp 0bp,clip=true]{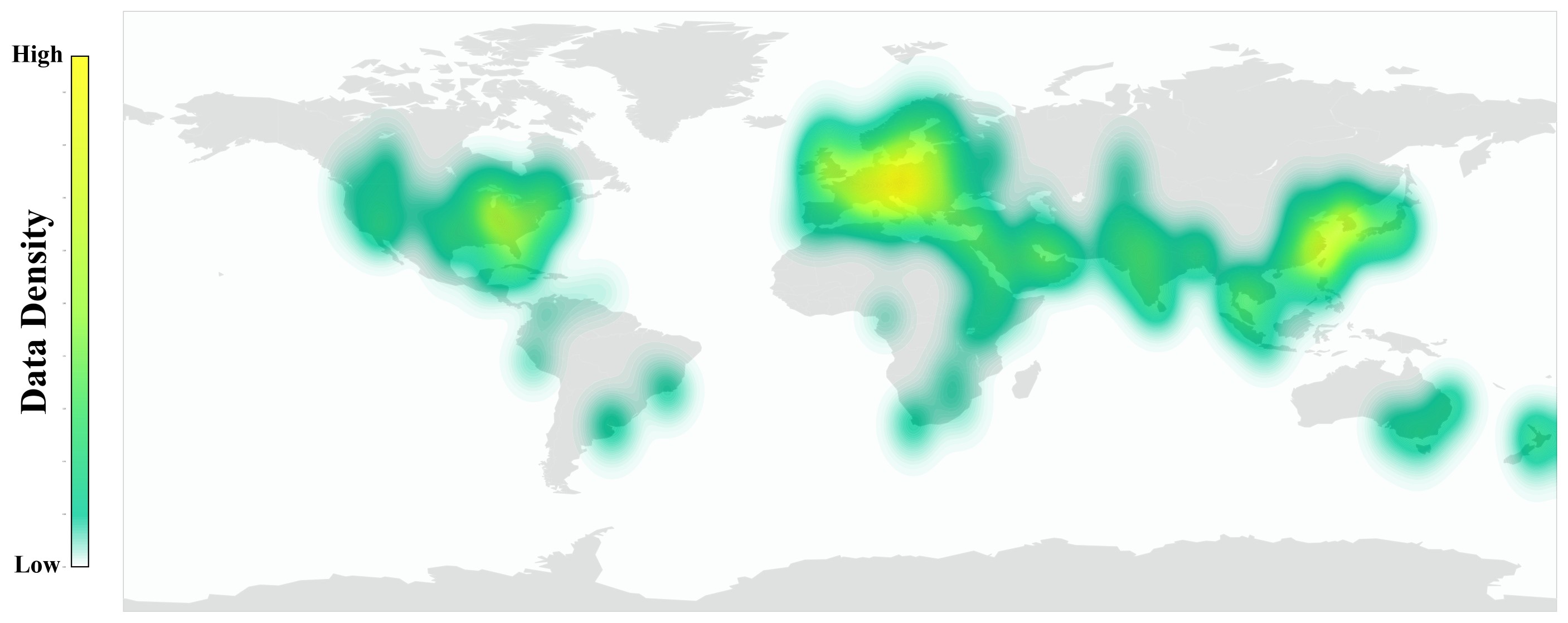}} 
	\caption{Data distribution. GeoGAT10k covers most regions of the world and maintains a uniform spread across the globe, ensuring balanced geographic diversity.}
	\label{fig-4} 
\end{figure}

Unlike CityGuessr68k~\cite{1}, which consists primarily of first-person driving or walking single-shot continuous videos, GeoGAT10k is sourced from TikTok and comprises multi-shot edited videos, each typically under one minute. Despite the short duration, the editing introduces richer geographic cues within a single video, imposing greater demands on spatiotemporal modeling. For preprocessing, we uniformly sample 50 frames per video and assign hierarchies, where the city serves as the finest granularity, from which the corresponding state, country, and continent are derived. GeoGAT10k contains 9,720 videos (486,000 frames) spanning 166 cities, 157 states/provinces, 91 countries, and 6 continents, with detailed distribution shown in Fig.~\ref{fig-4}. We further evaluate generalization across different video types on GeoGAT10k in Section~\ref{sec:geogat10k}. 

\textbf{CityGuessr68k dataset.} To comprehensively evaluate the effectiveness of GeoGAT, we conduct ablation and comparative experiments primarily on the public CityGuessr68k~\cite{1}. Following the setup of CityGuessr~\cite{1}, we stratify the dataset into an 80:20 train–test split, preserving class distributions across the four geographic hierarchies (city, state, country, continent). This yields 54,614 training videos and 13,655 test videos.

\textbf{Metrics.} We employ Top1 accuracy for each hierarchy as the primary evaluation metric, assessing the model’s localization performance across four hierarchies: city, state/province, country, and continent. We show additional analysis on Top3/Top5 accuracy on CityGuessr68k and GeoGAT10k in the supplementary material.

\subsection{Training Details}
\label{sec:training}

We implement GeoGAT in PyTorch~\cite{33}. For each video, $m=16$ frames are sampled bidirectionally and resized to $224 \times 224$. We adopt VideoMAE-Small~\cite{29} pretrained on Kinetics-400~\cite{30} as the video encoder with feature dimension $d=384$, and set the offset $p=1$. The hierarchy graph is a 2-layer GATv2 with 4 attention heads per layer and dropout rate 0.2. The model is trained on a single NVIDIA RTX A6000 GPU with batch size 12, using AdamW~\cite{AdamW} at learning rate $5 \times 10^{-5}$. The temperature $\tau=2.3$ and loss balancing factor $\lambda=0.1$. CityGuessr68k is trained for 10 epochs and GeoGAT10k for 20 epochs; all other hyperparameters are identical across datasets. The predefined geographical hierarchy graph is provided in the supplementary material.

\begin{table}[t]
	\centering
	\footnotesize
	\setlength{\tabcolsep}{4.5pt}
	\begin{tabular}{@{}lc cccc@{}}
		\toprule
		Method & Venue & City & State & Country & Continent \\
		\midrule
		\multicolumn{6}{@{}l@{}}{\textit{Image-based}} \\
		PlaNet & ECCV'16 & 55.8 & 56.3 & 60.8 & 74.1 \\
		ISNs & ECCV'18 & 59.5 & 59.9 & 64.1 & 75.9 \\
		GeoDecoder & CVPR'23 & 64.2 & 64.5 & 69.5 & 79.9 \\
		GeoCLIP & NeurIPS'23 & 57.8 & 60.5 & 75.9 & 90.8 \\
		GeoReasoner & ICML'24 & 38.5 & 42.8 & 64.4 & 81.9 \\
		GeoBayes$^*$ & AAAI'26 & 45.3 & 53.1 & 78.9 & 91.1 \\
		\midrule
		\multicolumn{6}{@{}l@{}}{\textit{Video-based}} \\
		Timesformer & ICML'21 & 60.9 & 61.4 & 66.1 & 78.4 \\
		VideoMAE & NeurIPS'22 & 64.5 & 64.5 & 65.9 & 74.4 \\
		CityGuessr & ECCV'24 & 69.6 & 70.2 & 74.8 & 83.8 \\
		VidTAG-cls$^\dagger$ & CVPR'26 & 79.3 & 81.0 & 85.4 & 91.8 \\
		VidTAG-ret$^\ddagger$ & CVPR'26 & \underline{81.9} & \underline{82.4} & \underline{85.8} & \underline{91.9} \\
		\midrule
		GeoGAT & \textit{Ours} & \textbf{84.5} & \textbf{84.9} & \textbf{86.8} & \textbf{92.2}\\
		\bottomrule
	\end{tabular}
	\caption{Comparison with state-of-the-art methods. VidTAG-cls$^\dagger$: visual encoder adapted as a classifier under our training protocol. VidTAG-ret$^\ddagger$: native retrieval paradigm, $k$-NN ($k=5$) over training embeddings without GPS queries. GeoBayes$^*$ is reproduced following the original paper. Best baseline results are \underline{underlined}; best overall results are in \textbf{bold}. Details are in the supplementary material.} 
	\label{tab:cityguessr68k}
\end{table}

\subsection{Performance on CityGuessr68k}
\label{sec:cityguessr68k}

Following prior works~\cite{1}, we compare GeoGAT with recent baselines on 
CityGuessr68k, including image-based methods 
(PlaNet~\cite{3}, ISNs~\cite{5}, GeoDecoder~\cite{27}, GeoCLIP~\cite{GeoCLIP}, 
GeoReasoner~\cite{GeoReasoner}, GeoBayes$^*$~\cite{GeoBayes}) and video-based 
ones (Timesformer~\cite{timesformer}, VideoMAE~\cite{29}, CityGuessr~\cite{1}, VidTAG~\cite{kulkarni2026vidtag}). 
Since VidTAG's retrieval paradigm requires GPS queries at inference, which 
are unavailable for geo-localization, we evaluate it under two fair protocols: 
VidTAG-cls$^\dagger$, which fine-tunes its visual encoder as a classifier 
under identical training conditions, and VidTAG-ret$^\ddagger$, which 
preserves its native retrieval strength via $k$-NN matching over training 
embeddings. 

Table~\ref{tab:cityguessr68k} reports the results. GeoGAT achieves the best 
performance at every granularity against both VidTAG variants, indicating 
that the advantage stems from modeling design rather than the comparison 
protocol. The gains are most pronounced at finer granularities, where 
single-frame evidence is ambiguous and our bidirectional temporal sampling 
aggregates complementary cues across the video. Notably, VidTAG-ret$^\ddagger$ 
outperforms its classifier-adapted counterpart, confirming that VidTAG's 
strength lies in retrieval; yet GeoGAT still surpasses it, demonstrating 
that graph-structured hierarchical classification captures fine-grained 
geographic evidence beyond either paradigm.

\subsection{Performance on GeoGAT10k}
\label{sec:geogat10k}

We evaluate GeoGAT on GeoGAT10k to verify generalization to multi-shot edited videos. The dataset is split into training and test subsets at a 75:25 ratio, with all hyperparameters consistent with those adopted for CityGuessr68k.

\begin{figure}[t]
	\centering
	{\includegraphics[width=0.97\linewidth,trim=0bp 0bp 0bp 0bp,clip=true]{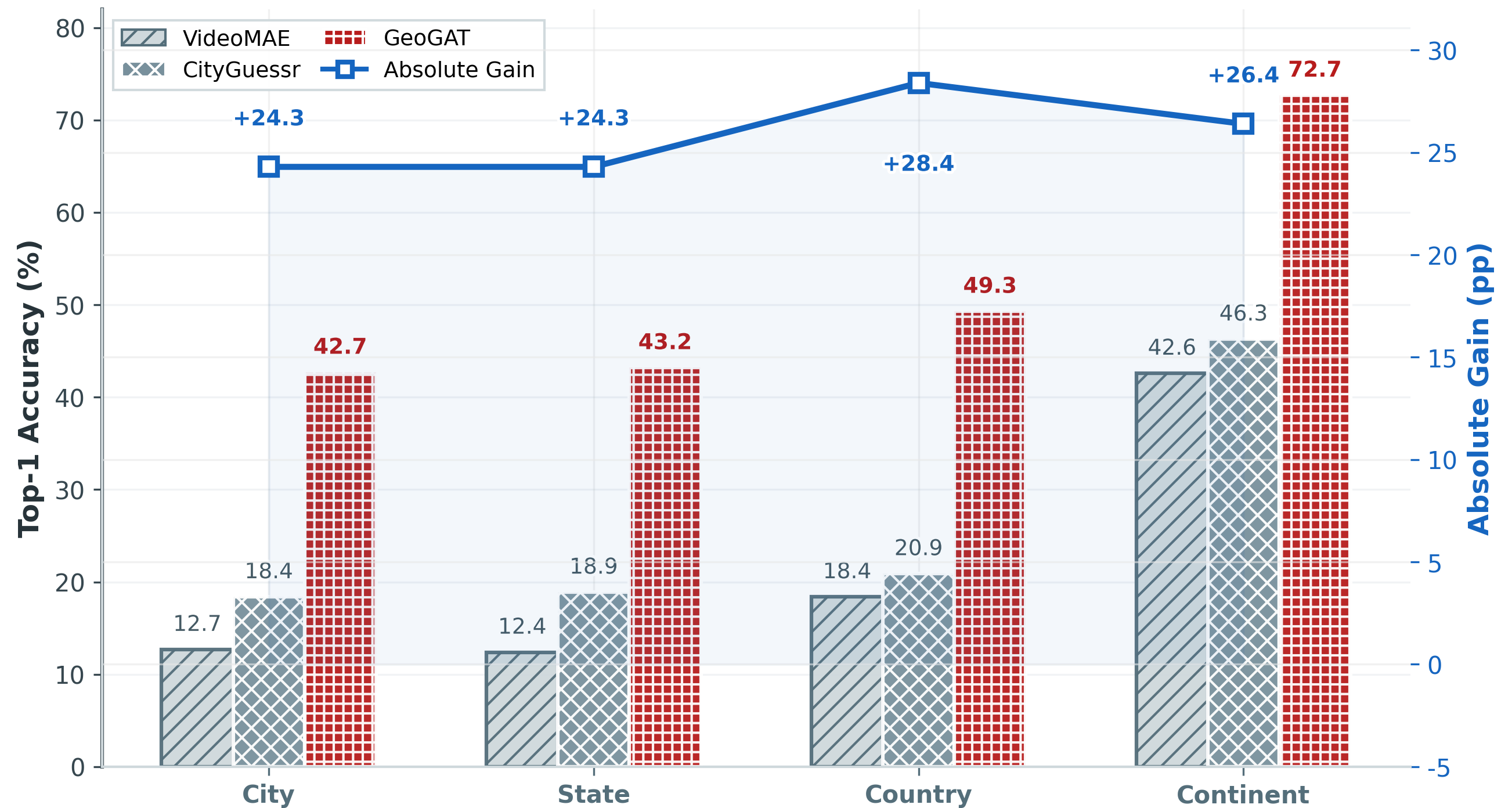}} 
	\caption{Top-1 accuracy comparison and gain of GeoGAT over the best baseline across geographic hierarchies.}
	\label{fig:geogat10k} 
\end{figure}

As shown in Fig.~\ref{fig:geogat10k}, GeoGAT achieves the best performance across all metrics. It outperforms CityGuessr by 24.3, 24.3, 28.4, and 26.4 percentage points at the city, state, country, and continent levels. All methods show lower accuracy on GeoGAT10k than on CityGuessr68k. This drop comes from a fundamental data difference: GeoGAT10k contains TikTok videos with abrupt scene transitions and visual effects, causing high-frequency discontinuities in geographic cues. CityGuessr68k uses continuous driving or walking footage with smooth geographic progression. This gap confirms GeoGAT10k as a harder benchmark for discontinuous scenarios. GeoGAT maintains strong performance on this dataset, validating its generalization to multi-shot edited videos.

\begin{table*}[t]
	\centering
	\footnotesize
	\setlength{\tabcolsep}{4.9pt}
	\begin{tabular}{@{}llccccc cccc cccc@{}}
		\toprule
		& & & & & & & \multicolumn{4}{c}{CityGuessr68k Top 1 Acc.(\%)} & \multicolumn{4}{c}{GeoGAT10k Top 1 Acc.(\%)} \\
		\cmidrule(lr){8-11} \cmidrule(lr){12-15}
		& Configuration & Backbone & Sampling & Soft & Hard & $\mathcal{L}_{\text{MI}}$ & City & State & Country & Continent & City & State & Country & Continent \\
		\midrule
		& Linear & Linear & OUS & $\circ$ & $\circ$ & $\circ$ & 61.9 & 62.2 & 64.1 & 73.4 & 12.7 & 12.4 & 18.4 & 42.6 \\
		& Linear + BTS & Linear & BTS & $\circ$ & $\circ$ & $\circ$ & 57.8 & 58.0 & 62.4 & 73.4 & 12.0 & 11.9 & 16.7 & 47.5 \\
		\midrule
		& w/ OUS & GATv2 & OUS & $\circ$ & $\circ$ & $\circ$ & 74.7 & 75.3 & 75.6 & 84.1 & 30.7 & 30.9 & 34.1 & 62.4 \\
		& w/ BRS & GATv2 & BRS & $\circ$ & $\circ$ & $\circ$ & 82.2 & 82.6 & 82.5 & 88.2 & 41.8 & 41.9 & 47.7 & 70.5 \\
		& w/ BTS & GATv2 & BTS & $\circ$ & $\circ$ & $\circ$ & 83.1 & 83.2 & 84.7 & 89.3 & 42.0 & 42.1 & 48.5 & 71.3 \\
		& w/ Soft & GATv2 & BTS & $\checkmark$ & $\circ$ & $\circ$ & 82.5 & 82.8 & 84.7 & 89.5 & 42.2 & 42.4 & 48.8 & 71.6 \\
		& w/ Soft + Hard & GATv2 & BTS & $\checkmark$ & $\checkmark$ & $\circ$ & 84.1 & 84.3 & 86.2 & 91.4 & 42.3 & 42.5 & 48.9 & 72.0 \\
		& GeoGAT & GATv2 & BTS & $\checkmark$ & $\checkmark$ & $\checkmark$ & \textbf{84.5} & \textbf{84.9} & \textbf{86.8} & \textbf{92.2} & \textbf{42.7} & \textbf{43.2} & \textbf{49.3} & \textbf{72.7} \\
		\bottomrule
	\end{tabular}
	\caption{Ablation study on CityGuessr68k and GeoGAT10k. $\circ$: disabled; $\checkmark$: enabled. OUS: one-way uniform sampling; BRS: bidirectional random sampling; BTS: bidirectional temporal sampling. The best results are in \textbf{bold}.}
	\label{tab:ablation}
\end{table*}

\subsection{Ablation Study}
\label{sec:ablation}

We conduct ablation experiments on CityGuessr68k and GeoGAT10k with eight configurations (Table~\ref{tab:ablation}) to isolate the contribution of each component.

\textbf{Hierarchical classifier.} Replacing the linear heads with the graph-based classifier brings substantial gains under identical one-way uniform sampling (OUS) on both datasets. Notably, equipping the linear baseline with bidirectional temporal sampling (BTS) instead degrades performance, indicating that linear models lack the capacity to fuse complementary temporal views. Structure-aware message passing is therefore essential for converting the enriched features into effective hierarchical predictions.

\textbf{Sampling strategy.} With the graph classifier fixed, BTS consistently outperforms both OUS and bidirectional random sampling (BRS) across all hierarchies on both datasets. The offset-reversed view captures sparse geographic cues that one-way sampling misses, while its fixed stride preserves the temporal regularity that random sampling breaks, which explains the advantage of ordered complementary coverage.

\textbf{Hierarchical constraints.} The soft constraint produces mixed effects: marginal improvements at coarser levels but degradation at finer granularities, as soft parent-child coupling introduces over-smoothing that harms fine-grained discrimination. In contrast, the hard constraint yields consistent gains across all hierarchies, since enforcing strict geographic validity prunes the prediction space without interfering with representation learning. Rule-based enforcement thus proves more reliable than soft probabilistic coupling.

\textbf{Mutual information loss.} Adding $\mathcal{L}_{\text{MI}}$ brings consistent improvements, particularly at coarser levels, and the full model attains the best performance across all metrics. This confirms that the contrastive bound and the cross-entropy objective cooperate as two complementary variational bounds, jointly maximizing the mutual information between video representations and geographic labels.

\begin{figure}[t]
	\centering
	\includegraphics[width=0.95\linewidth]{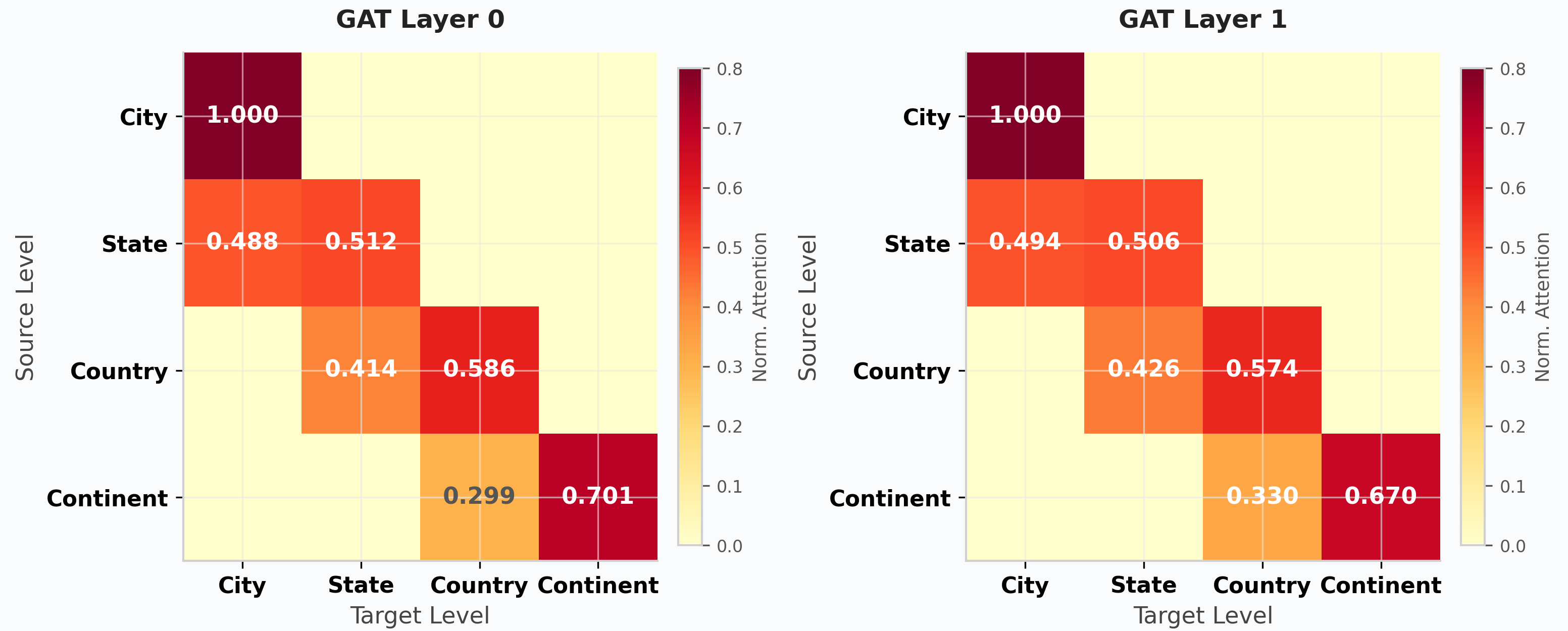}
	\caption{Normalized attention weight distribution across GAT layers. Cell $(i,j)$ shows attention proportion from source hierarchy $i$ (rows) to target $j$ (columns). Diagonal cells reflect self-attention, with dominance growing coarser from city to continent. Off-diagonal warmth along adjacent levels indicates cross-hierarchy flow, while distant cells remain cold, revealing locality in geographic dependencies.}
	\label{fig:attention}
\end{figure}

\subsection{Attention Flow Analysis}
\label{sec:attention}

To understand how geographic information propagates through the hierarchy graph, we visualize the learned attention weights in Fig.~\ref{fig:attention}. Each hierarchy primarily relies on its own features, with self-attention dominance growing from fine to coarse hierarchies. Cross-hierarchy attention concentrates along adjacent levels and decays with depth, reflecting the locality of geographic dependencies. This locality also explains the mixed effect of the soft constraint in Table~\ref{tab:ablation}: attention already captures first-order parent--child dependencies, so soft coupling adds redundancy and over-smoothing. This decay creates an information bottleneck at the continent level, which the hard constraint compensates for via rule-based validity enforcement. Consistent patterns across layers confirm stable structural properties, enabling reliable cross-granularity reasoning.

\subsection{Performance on Sparse Sample Regions}
\label{sec:sparse}

\begin{figure}[t]
	\centering
	\includegraphics[width=1\linewidth]{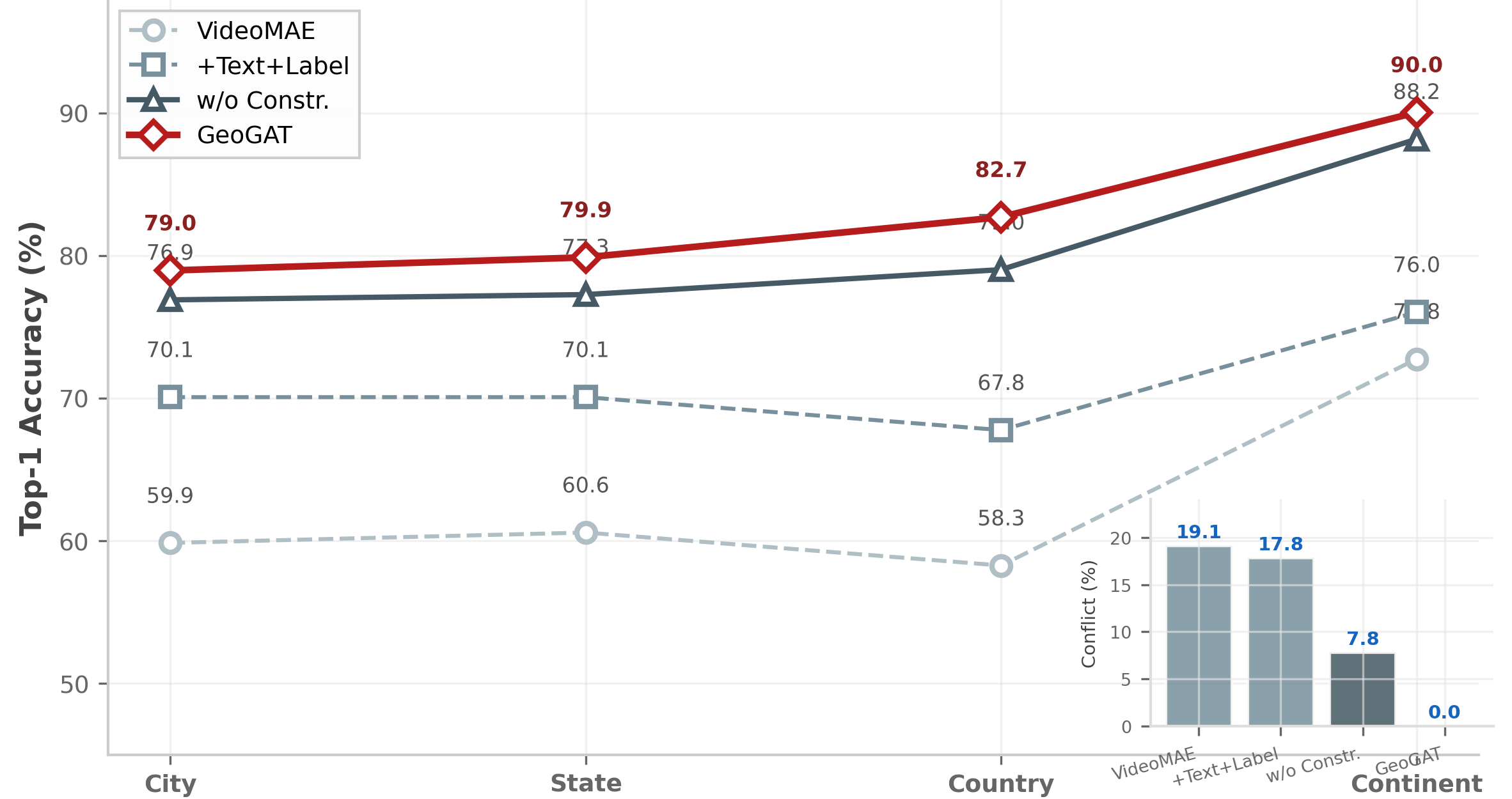}
	\caption{Sparse-sample region performance. Lines show Top-1 accuracy trends across geographic hierarchies; the right panel reports hierarchy conflict rates.}
	\label{fig:sparse}
\end{figure}

We construct a few-shot subset from CityGuessr68k containing 38 cities with fewer than 160 training samples and 822 test videos. Fig.~\ref{fig:sparse} compares GeoGAT with baselines on this subset. GeoGAT achieves the best performance across all hierarchies, with its advantage widening from city to continent. This widening is expected: rare cities share country- and continent-level labels with data-rich cities, so coarser hierarchies benefit most from graph-based knowledge propagation. For hierarchical consistency, the GeoGAT variant without constraints (w/o Constr.) already reduces the conflict rate to 7.8\%, compared with 17.8\% for CityGuessr and 19.1\% for VideoMAE, indicating that graph-structured message passing alone mitigates long-tail bias by propagating knowledge from data-rich to data-scarce labels. Adding both constraints further eliminates conflicts entirely, validating the dual-constraint design in sparse-sample regions.

\section{Conclusion}
\label{sec:conclusion}

We propose GeoGAT for global video geo-localization. GeoGAT couples bidirectional temporal sampling with graph attention networks and enforces hierarchical consistency through dual constraints. It achieves state-of-the-art accuracy across four geographic hierarchies on both CityGuessr68k and GeoGAT10k, outperforming the strongest baseline VidTAG, evaluated under both classification and retrieval protocols, by 2.6 percentage points at the city level. These gains stem from the effective fusion of complementary temporal views and structured geographic reasoning, which existing approaches lack. GeoGAT generalizes robustly to multi-shot edited videos where existing methods degrade substantially. We also release GeoGAT10k, a dataset of 9,720 videos from 166 cities featuring real-world editing patterns. The code and dataset will be released with the final version.

\clearpage

\bibliography{arxiv-geogat}

\end{document}